\RequirePackage{fix-cm}
\documentclass[smallextended]{svjour3}       
\smartqed  
\usepackage{graphicx}
\usepackage{hyperref}
\usepackage{amssymb, amsmath}
\usepackage{caption}
\usepackage[table,xcdraw]{xcolor}
\usepackage{xcolor}
\usepackage{ulem}
\newcommand{\emm}[1]{{\textcolor{blue}{#1}}}

\usepackage{amsmath}

\usepackage{wrapfig}
\usepackage{picins}

\begin{document}


\title{Evaluating sleep-stage classification: how age and early-late sleep affects classification performance}



\author{Eugenia Moris         \and
        Ignacio  Larrabide 
}


\institute{Eugenia Moris \at
                Universidad Nacional del Centro de la Provincia de Buenos Aires, Exactas,
                PLADEMA Institute, Yatiris Group, Tandil, Buenos Aires, Argentina.\\
                CONICET, Tandil, Buenos Aires, Argentina.\\
                Tel.: +54 0249 438-5690\\
                \email{emoris@pladema.exa.unicen.edu.ar}   \\
                The code used was uploaded to the git repository on \url{github.com/eugeniaMoris/2022_sleep_scoring_analisis}
           \and
           Ignacio Larrabide \at
                Universidad Nacional del Centro de la Provincia de Buenos Aires, Exactas,
                PLADEMA Institute, Yatiris Group, Tandil, Buenos Aires, Argentina.\\
                CONICET, Tandil, Buenos Aires, Argentina.\\
                Tel.: +54 0249 438-5690\\
}

\date{Received: date / Accepted: date}

\maketitle

\begin{abstract}
Sleep stage classification is a common method used by experts to monitor the quantity and quality of sleep in humans, but it is a time-consuming and labour-intensive task with high inter- and intra-observer variability. Using Wavelets for feature extraction and Random Forest for classification, an automatic sleep-stage classification method was sought and assessed. The age of the subjects, as well as the moment of sleep (early-night and late-night), were confronted to the performance of the classifier. From this study, we observed that these variables do affect the automatic model performance, improving the classification of some sleep stages and worsening others.


\end{abstract}
    
\par\noindent 
\parbox[t]{\linewidth}{
\noindent {\bf Eugenia Moris} is a Software Engineer and a PhD student in Computational and Industrial Mathematics, working in EEG signal processing with machine learning techniques with a PhD scholarship granted by CONICET.}
\vspace{4\baselineskip}

\par\noindent 
\parbox[t]{\linewidth}{
\noindent {\bf Ignacio Larrabide} graduated as Software Engineer. He pursued his DSc in Computational Modelling at LNCC, Petrópolis, Brasil. He currently leads the Yatiris research team at Pladema Institute.}
\vspace{4\baselineskip}

\section{Introduction}
\label{intro}
Polysomnography (PSG) collects physiological parameters during the night in order to analyze patients' sleep. This procedure uses electroencephalogram (EEG), electrooculogram (EOG), electromyogram (EMG), electrocardiogram, pulse oximetry, airflow and respiratory effort. Sleep staging is determined using information from EEG, EOG and EMG electrodes. Electrical activity of frontal, central and occipital brain regions, as well as eye movements and chin EMG are used to determine the sleep stage \cite{rundo2019polysomnography}.

PSG sleep scoring is a common method used by experts to monitor the quality and quantity of sleep in humans as well as for diagnosing sleep disorders \cite{mckinley2013sleep}. This procedure involves analysing an entire night's sleep segmented into 30-second epochs, each one of such epochs classified into a set of predefined sleep stages. Recent studies performed automatic sleep scoring using machine learning. The study of Silveira et al. \cite{da2017single} uses Random Forest (RF) for sleep scoring with discrete Wavelet as a feature extractor. The approach of Hassan et al.  \cite{hassan20171automated} classifies sleep stages based on a single EEG channel, while studies such as \cite{rahman2018sleep} classify them based on a single EOG channel. Further, RF is used for a series of EEG-related problems, such as early seizure detection \cite{donos2015early}, human mental state classification \cite{edla2018classification}, among others. Approaches like \cite{da2017single} use only EEG channel data. But further analysis, as presented by \cite{krakovska2011channels}, noticed an improvement in classification performance when using various types of channels. 

Considering the changes found in sleep during the natural ageing process in humans, we aim at analyzing how these changes could affect our classification model \cite{luca2015age,van2000age,moser2009sleep}, 
the article by Zhou et al. \cite{zhou2020automatic} employed RF and LightGBM as machine learning algorithms to classify sleep stages. Notably, they incorporated age as a feature in their classification task. On the other hand, papers like \cite{gais2000early,rasch2013sleep} analyse the difference between early sleep and late sleep, showing a large difference between them. In this paper, we also seek to analyse how these changes might affect the performance of the trained models.

Here we analysed a sleep scoring model to assess the effect of age, as well as the effect of early (early night, first 4 hours) and late (late night, last 4 hours) sleep, on the performance of a classical classifier on epoch sleep stage. For this purpose, we use wavelet as a feature extractor and RF as a classifier. As a result, we seek to understand how different models might best fit the needs of each problem. 

\section{Methods}
\label{sec:methods}

\subsection{Signals}
\label{sec:signals}

\begin{figure}[p]
    \centering
    \includegraphics[width=\linewidth]{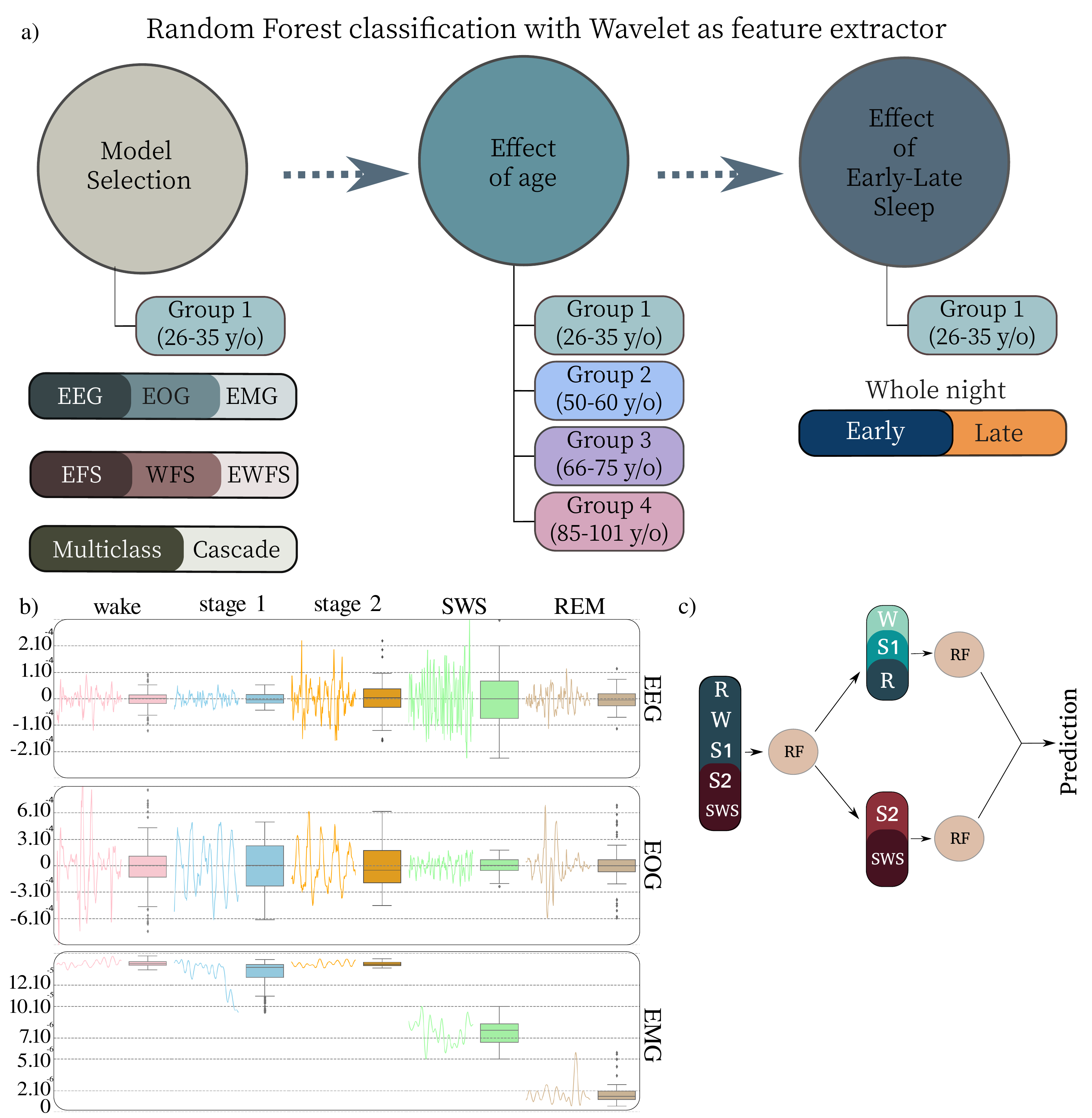}
    \caption{(a) We divided the experiments in three stages. In the first part, the model selection is done, where we choose the optimal model for our hypothesis and experiments. Here, we choose which channels, and whether a multi-class or a cascade model is used. In the second stage, we analyse how the classification performance can be affected by the subjects' age. In the third and final stage, we study the effect of early-late sleep. (b) Sample single epoch data for EEG, EOG and EMG channels. Signals are coloured differently for each sleep stage. Pink corresponds to wake (W) stage,  light-blue represents stage 1 (S1), correspond to stage 2 (S2), green correspond to slow-wave sleep (SWS) stage and, finally, brown indicates rapid eye movement stage (REM). On the right of each epoch, a box-plot shows the distribution of the signal. (c) Classification castade used for the cascade model (CM)} 
    \label{fig: fig1}
\end{figure}

Electroencephalography (EEG) is a popular non-invasive technique for recording electric signals generated by the action of brain cells using electrodes placed on the scalp, following the International 10–20 System \cite{jasper195810}. This method captures electrical activity in the cerebral cortex with high temporal resolution. Electrooculography (EOG) is used to record the eye movements of a stationary subject by placing electrodes at the canthi of the eyes \cite{blinowska2006electroencephalography}. Electromyography (EMG) measures muscle response or electrical activity in response to nerve stimulation of the muscle. For sleep scoring, a sub-mental chin EMG is usually employed. Figure \ref{fig: fig1} shows samples of 30 seconds for each channel and different sleep stages. Sleep-EDF dataset already segmented in 30 second epochs, which is a standard practice in EEG data analysis \cite{sharma2020automated,rahman2018sleep}.

\subsection{Sleep-EDF expanded} \label{sec:dataset}

We used the public dataset Sleep-EDF expanded (Sleep-EDFx) \cite{goldberger2000physiobank,kemp2000analysisSleep-EDF}, which has 153 files obtained between 1987-1991 in a study focused on the study of age effects on sleep in healthy Caucasians aged 25-101. In this article, all the subjects were not under any sleep-related medication \cite{mourtazaev1995age}. Two PSG of about 20 hours each were recorded during two subsequent day-night periods at the subjects' home. Each sample contains two EEG (from Fpz-Cz and Pz-Oz electrode locations), one EOG (horizontal), one submental chin EMG, and an event marker. All records were pre-classified by an expert following the R\&K criteria \cite{wolpert1969manual}. 

All experiments were run on Wake (W), REM, Stage 1 (S1), Stage 2 (S2) and Slow Wave Sleep (SWS). SWS results from merging Stages 3 and 4 according to R\&K criteria \cite{wolpert1969manual}. The EEG Fpz-Cz, EOG and EMG channels were used, the EOG and EEG signals were sampled at 100Hz; EMG was sampled at 1Hz \cite{goldberger2000physiobank} with a frequency response range (3dB points) from 0.5 to 100Hz \cite{mourtazaev1995age}. To maintain the recorded performance, a criterion was established where recordings were automatically deleted if the full-bandwidth (0.5-100 Hz) noise level of an EEG channel exceeded 7 IL V rms. Epochs marked by experts as movement time and not scored were excluded. PSG were recorded during the full day for all the subjects. W samples of the daytime were removed to avoid noisy information, maintaining W stage samples 20 minutes before and 20 minutes after sleep. As a pre-processing we only maintained the pre-processing made by the data-generators \cite{mourtazaev1995age} were they standardized the digital amplitudes using amplitude calibration signals and checked the noise levels using zero-voltage calibration. We do not apply any other pre-processing to the signal. A notch filter at 50 Hz was also applied on the data.

As we can observe in the table \ref{tab: age_groups}, the subjects were divided by their age into four groups. Group 1 (26 - 35 years old), Group 2 (50 - 60 years old), Group 3 (66 - 75 years old), and Group 4 (85 - 101 years old) following the work of \cite{mourtazaev1995age}.

\begin{table}[]
\centering
\captionsetup{justification=centering}
\caption{Subjects description}
\begin{tabular}{|c|c|c|}
\hline
Groups & Age range  & \#subjects \\ \hline
G1     & 26-35 y/o  & 19         \\ \hline
G2     & 50-60 y/o  & 20         \\ \hline
G3     & 66-75 y/o  & 19         \\ \hline
G4     & 85-101 y/o & 16         \\ \hline     
\end{tabular}
\label{tab: age_groups}

\end{table}

\subsection{Wavelet transform for feature extraction}
\label{sec:wavelet}

Wavelet transform (WT) is a powerful tool to extract features from non-uniform 1D signals, like EEG, EOG and EMG \cite{clerc2016brain}. Unlike the Fourier transform, the WT can characterise time information in addition to frequency information, achieving a good trade-off between temporal and frequency resolution. There are two types of wavelets: continuous and discrete. We used discrete wavelets since they are memory efficient and have proven to be highly effective in the analysis of sleep EEG \cite{daubechies1992ten}.

WT uses finite basis functions called wavelets, which are scaled and translated copies of a single finite length waveform known as the mother wavelet \cite{rao2013brain}. In this approach, we considered different wavelet's families, namely: Bior\-tho\-gonal ``Discrete" FIR approximation of Meyer Wavelet, Coiflets, Daubechies, Haar, Reverse biorthogonal and Symlets.

After applying the WT, detailed and approximation coefficients are obtained. Over the approximation coefficients, we apply WT again to extract more specific and local information. This repetition can be performed recursively as many times as desired. However, after several iterations, the model focuses on too small regions, only capturing noise instead of actual signal features. Here, we tested recursion levels 1, 2, 3, and 4.

We applied WT on each epoch (30 seconds), and the derived coefficients were considered features. 
\emm{Driven by a future objective of achieving real-time classification of sleep, we applied standardization to the training data-set. The mean and standard deviation used for standardization of training data were saved and used on validation and test data-sets. This decision was motivated by the impossibility of subject-specific standardization in online sleep stage classification.}

The use of statistical values extracted from the WT coefficients as features for each sample was also analysed. The wavelet coefficients' mean, median, standard deviation, kurtosis and skewness were calculated for each sample. Figure \ref{fig: fig1} summarises this information in box-plots next to each signal sample for the different sleep stages. 

\subsection{Random Forest non-linear classification}
\label{sec:random_forest}

RF is a classification algorithm consisting of a combination of tree predictors, where the nodes are split by a random selection of features \cite{breiman2001random}. 
The standardised feature-sample matrix is provided as input. Training, validation and test split of the sample data were done by subject. 70\% of the subjects were used for training and validation, and the remainder 30\% was saved for testing. During model training, we used the Leave-One-Out cross-validation method. In this method, one subject is separated and used to validate a model trained with the rest of the subjects. This step is repeated until all subjects have passed through the validation stage. Finally, the mean and standard deviation resulting from all validations is considered for selecting hyper-parameters.
Random seed '1234' was used to ensure that the same sequence of random numbers was generated in subsequent algorithm runs. To ensure we get the same randomness in the bootstrapping of the samples used when building trees and the same sampling of the features to consider when looking for the best split at each node.

\subsection{Cascade model}
\label{sec:cascade_model}

We hypothesize that if we separate the classes and make classifications over fewer samples, we could obtain more specific features and, thus, a better performance. Figure \ref{fig: fig1} shows the distribution of the different signals. We notice that, in EEG, the signal on W, S1 and REM are similar; meanwhile, S2 and SWS stages show a larger amplitude.

Figure \ref{fig: fig1} shows a schematic representation of the Cascade Model (CM). We divide the training into two levels. The first level divides the input into two groups. Group 1 contains epochs of W, S1 and REM. On the other hand, S2 and SWS stage was part of Group 2. Then, we train an SM model that classifies samples between these two groups. At the second level, two RF models are used to finalise the classification of subgroups generated in the previous level. One model produces a sub-classification for samples in Group 1, and another model classifies samples in Group 2 (S2 and SWS).

\subsection{Model setup and experiments}
\label{sec:model_setup}

A series of experiments were conducted to select the best configuration for the model setup and training.

First, a multi-class model was trained on the wavelet coefficients used as features, recalled as Wavelet Model (WM). We trained three WMs; one only used the EEG channel, a second used EEG and EOG channels and a third one trained with EEG, EOG and EMG channels. The model with the best performance was used for the classification.

Having selected the channel, we analysed the features to be used by the model. For this, we train and compare three different models again. WM, the model previously presented. We also considered a Statistical values Model (SM), a model that uses statistical values obtained from the wavelet coefficients (mean, median, standard deviation, kurtosis and skewness) as features. And finally, the Extended Model (EM), uses the WT coefficients as well as the statistical values as features.

As shown in Figure \ref{fig: fig1}, the signals on the distinct sleep stages had different distributions. Because of that, we hypothesise that the SM and the EM, the models using statistical values, will have good performance. 

Having selected the channels and the features, we compare the multi-class model with a CM. We hypothesise that using the CM, which classifies between fewer classes, can improve classification performance.  

In the second part of the work, we want to assess how the extensive range of ages between the subjects in the sample will affect classification performance. We know that the power of brain waves decreases with increasing age \cite{luca2015age}. For this reason, the subjects were separated into age groups to assess how it affects classification performance. We hypothesise that training our model with subjects of the same age as those to be classified will improve classification performance.

Finally, knowing that the EEG waves from the first half of the night (early sleep) show a discrepancy from those observed during the night's second half (late sleep) \cite{rasch2013sleep}, we studied how these affect sleep stage classification. We hypothesise that using early-stage samples for training, where characteristic waves are more robust and clearly defined, will improve classification performance.

For each of the aforementioned experiments, the models were trained using different sets of hyperparameters. The performance of different wavelet families was evaluated, including Biorthogonal ``Discrete" FIR approximation of Meyer Wavelet, Coiflets, Daube-chies, Haar, Reverse biorthogonal, and Symlets, to identify the one that yielded the best results. Additionally, the impact of varying the percentage of Random Undersampling (RUS) applied to the W stage was analyzed, ranging from 1\% to 100\% in increments of 5\%. The depth level of the wavelet was also considered, ranging from 1 to 5. Furthermore, Principal Component Analysis (PCA) was performed with varying numbers of final characteristics, ranging from 50 to 400 features in increments of 50. Lastly, the effect of changing the number of trees in the ensemble was investigated, ranging from 50 to 500 trees in increments of 50. These comprehensive experiments allowed for a thorough exploration of different hyper-parameter combinations and their impact on the model's performance. For each model, the set of hyper-parameters that produced the best results on the training-set, were used.

\subsection{Effect of age on sleep stage classification}
\label{sec:age}

Sleep efficiency, SWS and the spectral power densities in slow waves and fast spindles decrease with age, while theta-alpha and beta power in non-rapid eye movement sleep increased in the elder \cite{luca2015age}.

With this knowledge, we hypothesise that given the changes in the sleep stages with age, a model trained with the younger subjects could show poorer performance in classifying signals obtained in older subjects, and vice versa. 

To verify our hypothesis, four models were trained, considering data from each subject age group. In this experiment, we used the model selected from the \textit{Model selection} part (Section \ref{sec:model_setup}). The resulting models were tested in subjects unseen by the model during training. 

\subsection{Effect of early/late night sleep on classification}
\label{sec:early-late}

SWS is predominant during the early part and decreases in intensity and duration across the sleep period. In contrast, REM sleep becomes more intense and extensive towards the end of the sleep period \cite{rasch2013sleep}.

We hypothesise that the model trained with early sleep will have bad results in REM stage and improve its performance in the classification of the SWS stage. On the other hand, the model trained with late sleep samples should present a poorer performance in the classification of SWS while having a better performance in the classification of REM.

To verify our hypothesis, we separate a night of sleep into two halves. The first half of the night, which we call early sleep, and the second half of the night, which we call late sleep. Then, two models are trained following the process described in section \ref{sec:model_setup}, trained with samples from early sleep and late sleep, respectively. We finally tested early and late sleep stage classification in both models.

\subsection{Computational implementation}
\label{sec:computational}

All the data and experiments were prepared using Python language scripts, a powerful high-level open-source language with a clean syntax. Many data handling and processing libraries made this language a suitable choice for scientific development \cite{oliphant2007python}. MNE library for analysis and data preparation of EEG signal (V$0.20.7$) was used. MNE is an open-source Python package for exploring, visualising, and analysing human neurophysiological data. Each 30-second epoch from the data-set was stored in the Epoch class, which made its manipulation and handling easier \cite{gramfort2013meg}. For wavelet applications, the Pywavelet library (V$1.1.1$) was used, which contains many discrete and continuous wavelets \cite{lee2019pywavelets}. For data analysis, Scikit-learn, a simple and efficient toolkit for predictive analysis (V$0.22.1$), was used. In particular, we use RF and Metrics methods \cite{scikit-learn}.

\subsection{Metrics and Validation}
\label{sec: metrics}

Different discrete wavelet families, provided by the PyWavelet library, were first analysed. Next, features were extracted at different scales from the signal by applying the WT recursively. Given that the number of features extracted using this multi-scale wavelet approach was high on WM and EM (3.000 features), PCA was used for dimensional reduction. This helped balance the number of features to the number of epochs. A reduction of features, ranging from 50 to 400, was tested at a step of 50. Finally, the last meta-parameter selected was the number of decision trees in the RF, testing between 50 to 500 trees steps of 50.

The optimal model's hyper-parameters were obtained using Leave-One-Out cross-entropy. 70\%-30\% of the subjects were set apart for train and validation, respectively. For validation, one subject from the train-validation set is separated, stage classification is done by the model and performance metrics are obtained. We repeat this process for all the subjects. We choose the final meta-parameter values using F-score as the quality measure, which supports best classification results in general \cite{sokolova2009systematic}.

For the multi-class classification, we utilized the F-score macro-averaging method, as suggested by Sokolova and Lapalme \cite{sokolova2009systematic}. The F-score macro-averaging method treats all classes equally when calculating the overall F-score. This approach ensures that each class contributes equally to the final evaluation, regardless of imbalances in the dataset. By considering the precision and recall for each individual class and then averaging them, the F-score macro provides a comprehensive assessment of the model's performance across all classes. This enables a fair evaluation, particularly when dealing with imbalanced class distributions.

Additional experiments were conducted to explore the impact of different machine learning methodologies while using the same training data-set. RF, Support Vector Machine (SVM), and k-Nearest Neighbors (KNN) were compared. The statistical variables of wavelet coefficients extracted from the EEG, EOG, and EMG channels were utilized as features. The predictions were generated on a dedicated validation data-set, allowing for a comprehensive evaluation of each methodology's performance. Upon evaluation, it was observed that the RF model outperformed the other methodologies in terms of classification performance. Therefore, based on these results, the RF approach was chosen as the preferred methodology for the classification task.








\subsection{Statistical assessment of features}
\label{sec:static}

To assess the capacity of the selected features to characterise the different classes, statistical tests were used. Shapiro-Wilk test was used to verify the normality of the sample, which was the case for the data (p$<$0.05). Further, following the study of \cite{hassan20171automated}, a one-way analysis of variance (ANOVA) was used to verify, statistically, that the selected characteristics could discriminate between the five classes. We considered a confidence of 95\%. Therefore a p-value of less than 0.05 would indicate that the variance of the characteristics differs significantly between the classes. Scipy library (V 1.4.1) was used for this assessment.

\section{Results} \label{sec:result}

\subsection{Model selection}

Table \ref{tab:channel_comparison} shows the F-score resulting from WM when we train with EEG alone; with EEG and EOG; and with EEG, EOG and EMG channels. The highest F-score is highlighted in bold.

Except for S1, all the classes got the best F-score in the model trained with the three channels. For S1, the model trained with two channels rendered the best performance. We observe that the largest difference corresponds to stage SWS, for the models trained with two and three channels, by $2.87$. 

\begin{table}[h]
\small
\centering
\begin{tabular}{|cccc|}
\hline
\multicolumn{1}{|c|}{F-score} & \multicolumn{1}{c|}{EEG} & \multicolumn{1}{c|}{EEG - EOG} & EEG - EOG - EMG \\ \hline
Global                      & 44.93                    & 45.10                          & \textbf{45.76}  \\ \hline
W                         & 57.23                    & 58.08                          & \textbf{58.22}  \\ \hline

S1                      & 2.35                     & \textbf{3.63}                  & 2.96            \\ \hline
S2                      & 69.77                    & 69.87                          & \textbf{70.41}  \\ \hline
SWS              & 51.26                    & 48.75                          & \textbf{51.62}  \\ \hline
REM                          & 44.06                    & 45.17                          & \textbf{45.57}  \\ \hline
\end{tabular}
\caption{F-score of the WM when different channels are used for training. Global F-score and F-score for each class. Best scores are highlighted in bold.}
\label{tab:channel_comparison}
\end{table}

\begin{figure}[p]

\includegraphics[clip,width=\columnwidth]{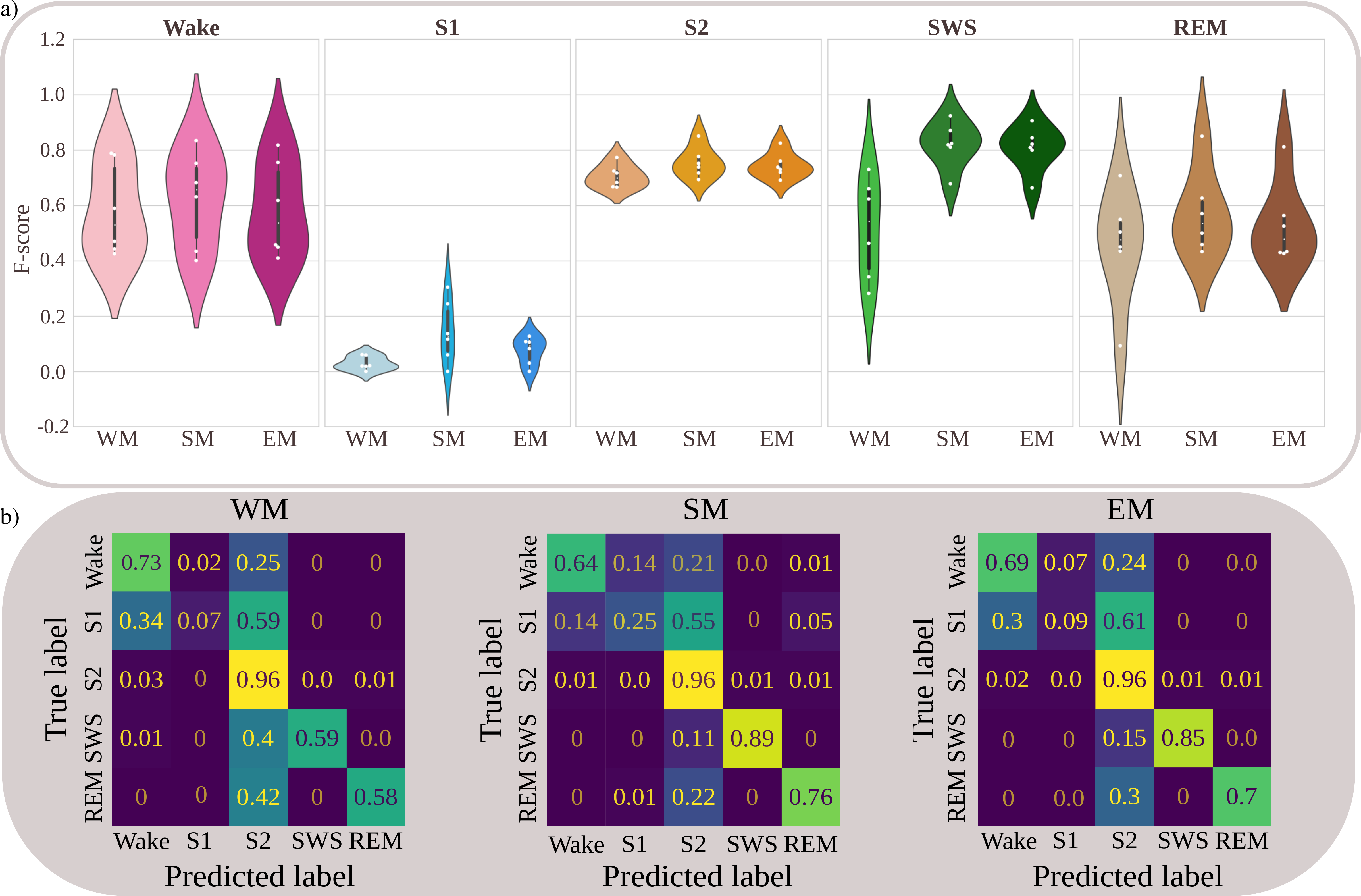}%

\caption{Results of the WM, SM and EM. (a) Violin graph of F-scores obtains for each class in the Wavelet Model (WM), in the Statistics variable Model (SM) and the Extend Model (EM). Each white dot represents the F-score obtained for each test subject. (b) Confusion matrix of the subject 14 obtained in each model}
\label{fig: Fig2}
\end{figure}

Figure \ref{fig: Fig2} shows the F-score obtained for the WM, SM and EM models presented in Section \ref{sec:model_setup}. 

F-score on the W stage, marked in pink, shows similar results in the three models, with significant variability in the results for each test subject. SM shows the minor subject with a low F-score compared to the other models. S1, indicated in blue, shows poor performance in general. We notice a difference in the performance of the SM model. EM shows a high result compared to the WM model. In S2, marked in yellow, we obtained similar results in all the models, with slightly higher results for SM and EM. Almost all the subjects obtained an F-score between 60\% and 80\%, where SM and EM improved that performance for one subject. There was one subject in the three models that obtained a higher F-score than the rest. 

Figure \ref{fig: Fig2} shows the confusion matrix obtained by each model for subject 14. For the three models, S2 had a correct classification in 96\% of the epochs. WM had a significant number of epochs correctly classified as W stage, but the model is mistaken 25\% of the time with S2. The SM model had a significantly correct classification of W stage in 64\%; nevertheless, the model confounded W epochs with S1 and S2. During the classification of S1, 25\% of the epochs got a correct classification. 

EM model got a correct classification of 69\% in W stage, but 24\% of the classification was wrongly classified as S2. S1, like the WM model, does not have high accuracy, while the model mistakes S1 with S2 in 61\% of the cases and with W stage in 30\%. 

\begin{table}[h]
\centering
\begin{tabular}{c|c|c|}
\cline{2-3}
                              & SM               & CM               \\ \hline
\multicolumn{1}{|c|}{Global} & 58.22\%          & \textbf{59.21\%} \\ \hline
\multicolumn{1}{|c|}{W}    & \textbf{62.16\%} & 60.01\%          \\ \hline
\multicolumn{1}{|c|}{S1} & 14.31\%          & \textbf{19.08\%} \\ \hline
\multicolumn{1}{|c|}{S2} & 75.39\%          & \textbf{76.86\%} \\ \hline
\multicolumn{1}{|c|}{SWS}     & 81.99\%          & 81.98\%          \\ \hline
\multicolumn{1}{|c|}{REM}     & 57.22\%          & \textbf{58.13\%} \\ \hline
\end{tabular}
\caption{F-score global and for each class, the higher results between models are marked in bold.}
\label{tab: CMvsSM}
\end{table}

Table \ref{tab: CMvsSM} shows the F-score obtained for each stage in the SM model and the CM. In general, CM model obtained a better result. In the W stage, the SM model shows a higher F-score than the CM model. In SWS classification, both models got the same metric. For the rest of the classes, the CM model improves the classification compared to SM model. 

    


\emm{The one-way ANOVA statistical test revealed significant differences between the five stages of sleep. The statistical variables (mean, median, std, kurtosis, skewness) obtained from the three channels were used as features.}

\begin{table}[]
\centering
\caption{Fscore from RF, SVM, and KNN algorithms prediction over the same validation dataset. The statistical variables of wavelet coefficients from three channels (EEG, EOG and EMG) were used as input, with default parameter values for each algorithm.}
\begin{tabular}{c|c|c|c|c|c|}
\cline{2-6}
                          & W    & S1   & S2   & SWS  & REM  \\ \hline
\multicolumn{1}{|c|}{SVM} & 0.75 & 0.36 & 0.83 & 0.86 & 0.75 \\ \hline
\multicolumn{1}{|c|}{\textbf{RF}}  & \textbf{0.78} & \textbf{0.37} & \textbf{0.84} & \textbf{0.87} & \textbf{0.77} \\ \hline
\multicolumn{1}{|c|}{KNN} & 0.71 & 0.30 & 0.80 & 0.83 & 0.71 \\ \hline
\end{tabular}
\label{tab:metodology_comparison}
\end{table}

The F-scores obtained by each methodology for each sleep stage can be observed in Table \ref{tab:metodology_comparison}. RF consistently outperformed the other two methods across all classes. SVM closely followed, with relatively similar values, while KNN performed the least effectively among the three methodologies. This clear performance distinction emphasizes the strength of RF for sleep stage classification.

\subsection{Effect of age on classification}

\begin{figure}[p]
    \centering
    \includegraphics[width=\linewidth]{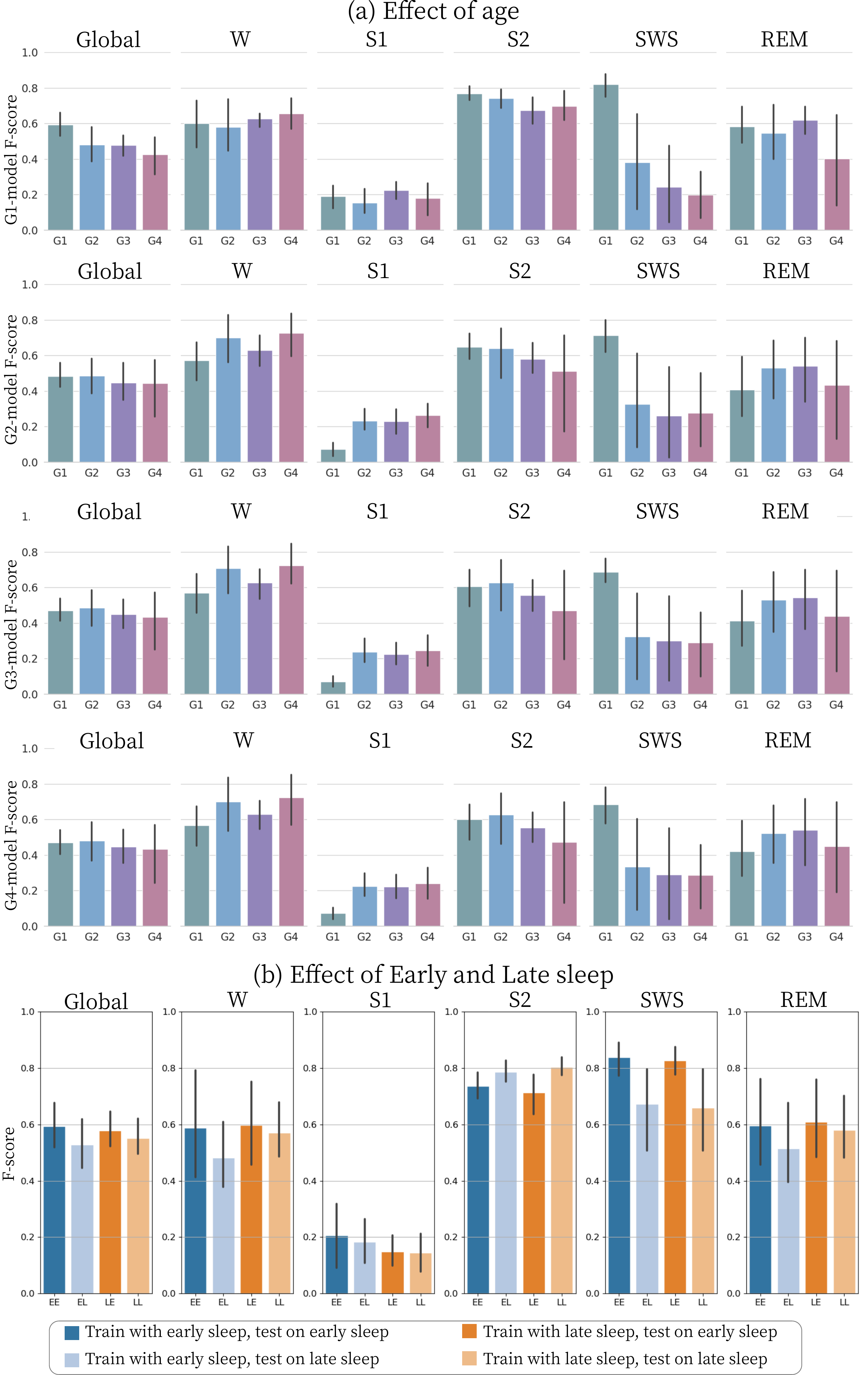}
    \caption{(a) F-score of four CMs, each one trained with different subject groups (G1, G2, G3 and G4), when classified over different groups. Groups are specified in Section \ref{sec:dataset}. Global F-score and the F-score for each class as well are shown. On the left side of each plot, we can observe the group subject used to train the model. We will call them G1-model, G2-model, G3-model and G4-model, based on the training subjects. (b) Global F-score and for each sleep stage. In blue, we present the result obtained with the model trained with early sleep. In orange, we show the results obtained with the models trained with late sleep. On the X-axis,  values are labeled using the samples used for training (first character, early or late), and the epochs was for testing (second character, again, early or late).}
    \label{fig: Fig3}
\end{figure}

Figure \ref{fig: Fig3} shows the result of four CMs trained with subjects of different age groups and then classified over the different test subjects from diverse age groups (Section \ref{sec:dataset}). We name the models based on the age of the subjects used for training (namely, G1-model, G2-model, G3-model and G4-model).

Generally, good results are obtained when G1-model is used to classify G1 subjects. G1-model improved the classification performance of G1 subjects over the rest. Further, this model is the one with a greater F-score for the classification of G3 subjects, in general. In the W stage, we notice that G1-model got lower classification results than the other models. In this model, we got a larger F-score when older subjects were classified. Results for S1, in general, were low for each model and each test subject. G1-model demonstrated a great result in S1 when G3 subjects were classified. 

For S2, contrary to S1, when older subjects were tested, a lower F-score was observed. G1-model shows higher results for all the test subjects than the other models. For REM stage, in almost every model, G2 and G3 subjects got the best result.

Based on the input to be classified, table \ref{tab:best_age} shows the models that performed the best for each stage.

\begin{table}[h]
\centering
\caption{The first column shows the input to classify. The model that best classifies global and for each stage depending on the input are marked in the Table with their corresponding colour. The model trained with G1 subject (G1-M) is marked in green, and the model trained with G2 subject (G2-M) is marked in blue. The model trained with G3 subject (G3-M) is marked in purple, and the model trained with G4 subject (G4-M) is marked in pink.}
\begin{tabular}{c|cccccc}
Input & Global                                                 & W                                                    & S1                                                 & S2                                                 & SWS                                                     & REM                                                     \\ \hline
Group 1     & \cellcolor[HTML]{76A5AF}{\color[HTML]{000000} G1-M} & \cellcolor[HTML]{76A5AF}{\color[HTML]{000000} G1-M} & \cellcolor[HTML]{76A5AF}{\color[HTML]{000000} G1-M} & \cellcolor[HTML]{76A5AF}{\color[HTML]{000000} G1-M} & \cellcolor[HTML]{76A5AF}{\color[HTML]{000000} G1-M} & \cellcolor[HTML]{76A5AF}{\color[HTML]{000000} G1-M} \\
Group 2     & \cellcolor[HTML]{C27BA0}G4-M                        & \cellcolor[HTML]{8E7CC3}{\color[HTML]{000000} G3-M} & \cellcolor[HTML]{6FA8DC}{\color[HTML]{000000} G2-M} & \cellcolor[HTML]{76A5AF}{\color[HTML]{000000} G1-M} & \cellcolor[HTML]{76A5AF}{\color[HTML]{000000} G1-M} & \cellcolor[HTML]{76A5AF}{\color[HTML]{000000} G1-M} \\
Group 3     & \cellcolor[HTML]{76A5AF}{\color[HTML]{000000} G1-M} & \cellcolor[HTML]{C27BA0}G4-M                        & \cellcolor[HTML]{6FA8DC}{\color[HTML]{000000} G2-M} & \cellcolor[HTML]{76A5AF}{\color[HTML]{000000} G1-M} & \cellcolor[HTML]{C27BA0}G4-M                        & \cellcolor[HTML]{76A5AF}{\color[HTML]{000000} G1-M} \\
Group 4     & \cellcolor[HTML]{C27BA0}G4-M                        & \cellcolor[HTML]{8E7CC3}{\color[HTML]{000000} G3-M} & \cellcolor[HTML]{C27BA0}G4-M                        & \cellcolor[HTML]{76A5AF}{\color[HTML]{000000} G1-M} & \cellcolor[HTML]{C27BA0}G4-M                        & \cellcolor[HTML]{C27BA0}G4-M                       
\end{tabular}
\label{tab:best_age}
\end{table}

\subsection{Effect of early-late sleep}

Figure \ref{fig: Fig3} shows the overall results for each sleep stage for the models trained with early or late sleep when they classify over early-late sleep in test subjects.

In general, the models' performance was better when classifying early than late sleep. The model trained with early sleep obtained the best result in classifying early sleep. However, this model shows a decay when testing over late sleep. For the W stage, almost all the results were similar. Only the classification of late sleep by the model trained with early sleep got a considerable decay in the classification. The classification of early W sleep was similar in both models. 

In the case of S1, the model trained with early sleep data obtained better results. 
The model trained with late sleep obtained a slight improvement in the classification of early sleep than late sleep. For S2, the models improved the classification of late sleep compared to the classification of early sleep. In classifying early sleep, the model trained with early sleep rendered better results than the model trained with late sleep. 

In REM stage, both models improved in the classification of early sleep. In the classification of REM in late sleep, the model trained with late sleep outperformed the model trained with early sleep.

In Figure \ref{fig: Fig3}, we can notice that there is low variability in the classification of all the test subjects. The model trained with late sleep shows more variability in the classification of early sleep. SWS stage shows considerable variability in both models when classified over late sleep. 

Table \ref{tab:best_earlyl} shows the best model depending on the input. We can observe that the late model got the best results in W and REM meanwhile early model got the best classification for S1 and SWS independently of the input.

\begin{table}[h]
\centering
\caption{The model that best classifies globally and for each stage depending on the input (early sleep or late sleep). The model trained with early sleep (Early-M) is marked in blue, and the model trained with late sleep (Late-M) is marked in orange}
\begin{tabular}{c|cccccc}
Input & Global                                                & W                                                   & S1                                                & S2                                                & SWS                                                    & REM                                                    \\ \hline
Early & \cellcolor[HTML]{3274a1}{\color[HTML]{000000} Early-M} & \cellcolor[HTML]{e1812c}{\color[HTML]{000000} Late-M} & \cellcolor[HTML]{3274a1}{\color[HTML]{000000} Early-M} & \cellcolor[HTML]{3274a1}{\color[HTML]{000000} Early-M} & \cellcolor[HTML]{3274a1}{\color[HTML]{000000} Early-M} & \cellcolor[HTML]{e1812c}{\color[HTML]{000000} Late-M} \\
Late  & \cellcolor[HTML]{e1812c}Late-M                        & \cellcolor[HTML]{e1812c}{\color[HTML]{000000} Late-M} & \cellcolor[HTML]{3274a1}{\color[HTML]{000000} Early-M} & \cellcolor[HTML]{e1812c}{\color[HTML]{000000} Late-M} & \cellcolor[HTML]{3274a1}{\color[HTML]{000000} Early-M} & \cellcolor[HTML]{e1812c}{\color[HTML]{000000} Late-M}
\end{tabular}
\label{tab:best_earlyl}
\end{table}

\section{Discussion} \label{sec: discussion}
In this approach, we study the variables to be considered when training an EEG sleep stage classification model. Specifically when using data from early and late sleep and data from subjects of different ages. The main contribution of this work is to the quantification of how these variables affect the accuracy performance of the derived model.

 We conducted an extensive analysis to understand how different models can classify the different sleep stages. On the one hand, we studied how adding different channels to an automatic classification model affected its performance. Also, we analysed whether the model performance improved when considering the raw wavelet coefficients, its statistical descriptors, or both. Finally, we considered using a cascade (multiple sequential binary classifications) or a single-pass multi-class RF model for the classification.

In a second experiment (Section \ref{sec:age}), we analysed the effect of subject age on both, model training and classification performance. We analysed how sleep varies in the classification with the age of the subjects. This analysis used a large data-set (80 subjects) segmented in different age groups.

In Section \ref{sec:early-late}, trained models were examined to assess their responses to early versus late sleep. There were structural changes in sleep stages depending on which half of the night sleep was considered, with certain epoch stages being increased or decreased. Other studies that manually analysed these changes found similar results \cite{rasch2013sleep}.

The discussion and analysis of the results were divided in three parts. First, we obtained a model to classify different sleep stages. In the second part, we analyse the effect of subjects' age in the classification. Finally, we study the effect of early and late sleep epochs on classification. 

\subsection{Model Selection}

An improvement in the classification performance was observed when the three channels were used for all the classes. The result is coherent since the more channels we use, the more information we provide to the model during the learning process. S1 showed the best classification results for the model trained with only two channels (EEG and EOG). This might be due to the similarity of EMG between W, S1 and S2 (Figure \ref{fig: fig1}). As we can see in the confusion matrix (Figure \ref{fig: Fig2}), the models frequently confuse S1 with either W or S2. So, using EMG information might prejudice S1, the class with fewer samples. SWS classification performance decayed in the model trained with two channels only, meaning that adding EOG is a confounding factor to the model, and it disfavours the classification of this class. 

The ANOVA statistical test showed that the selected features could discriminate between five classes with statistical significance. Observing that if the experiments are conducted in the correct order and making decisions based on a confidence metric, it is possible to select features that allow the correct discrimination between classes.

Krakovsk\'a and Mezeiov\'a analyse the best set of characteristics of PSG signals for the automatic classification of sleep stages \cite{krakovska2011channels}. From their results, the best combination of signals involved EEG, EOG and EMG channels to sleep scoring. The authors noticed that the EEG does not reveal much about the difference between S1 and REM. However, several EMG measures can be successful in the separation of W-REM stages and S1-REM. Further, the authors noticed that EOG channel helped differentiate between W and S1.

We observed an improvement in classification performance by the models trained with the statistical values, namely SM and EM, presented in section \ref{sec:model_setup}. These two models obtained an improvement in the classification performance of the SWS and REM stages (Figure \ref{fig: Fig2}). The difference in EMG signal variability of these two classes may explain the improvement (Figure \ref{fig: fig1}) due to the fact that, instead of using many unnecessary features, we focused on the features that best differentiate between classes, as observed in the past \cite{da2017single,sharma2018accurate}. 

All models showed difficulties in classifying S1, confounding it with W and S2 stages. WM was the model with the most errors between these classes. EM reduced this problem, and SM improved the classification of S1, reducing the errors, especially with the W stage. We speculate that the improvement of SM is due to the fact that the features better describe the difference between these three classes. 

The approach of Sors et al. \cite{sors2018convolutional} showed similar performance in the classification of sleep scoring. In their study, SWS was mistaken with S2, which also happens in our WM model. Further, Sors et al. mistakenly classified S1 as W, S2 and REM, which was also observed in our experiments. Finally, the authors show miss-classifications between REM, S2 and W, which we also observed. Unlike Sors et al., which only uses the EEG channel, fewer mistakes were found in the REM stage for our results. This has been observed in the past, where adding EOG and EMG channels helps the model to differentiate REM from the other classes \cite{krakovska2011channels}.

When using a CM, although the difference was small, we notice an improvement in almost every class except for the W stage. A possible explanation for the improvement is the separation of the three most similar classes, in the EEG channel, from the other two (Figure \ref{fig: fig1}). We notice that S1 is the class with the most considerable improvement when using the CM instead of SM. Similarly, Supratak et al. \cite{supratak2017deepsleepnet} noticed that in representation learning, the first filters on the convolutional layers separate the classes into two groups, W-S1-REM and S2-SWS, demonstrating similarities between the classes. Both models were compared, considering the optimal hyperparameter adjustments achieved during training. Each model was independently fine-tuned to obtain its best version before the comparison took place. However, since the differences in values across the S2, SWS, and REM stages are small, a different conclusion could arise from new experiments.



To avoid the adverse effects of data imbalance, the study of \cite{hassan20171automated} used Random Undersampling boosting (RUSboots) as a classifier, obtaining an accuracy for five classes of $83.49\%$. On the other hand, in our approach, we separately used the RUS method to reduce the imbalance of the data and an RF for classification. We obtained an overall hit of $82.76\%$ for five classes. We can observe that for both papers, the results were similar. However, keeping the control of the data imbalance and the classifier separate allows for adjusting the parameters of each method more precisely. It is important to note that the RUS technique was only applied to the W stage. This decision was based on the fact that the W stage showed the greatest difference compared to other sleep stages. The second stage with the most significant difference in the number of epochs was the S1 stage. However, unlike the W stage, the S1 stage had fewer epochs compared to the other stages. To balance this stage without losing information from the other stages, it would be necessary to increase the number of epochs in the S1 stage with more data.

\subsection{Effect of age}

Table \ref{tab:best_age} shows, for each stage, the model that produced optimal results as a function of the input. We observe that in classifying subjects of different age groups, the G1-model performed best, appearing most of the times in the Table. In some sleep stages, the G4-model, however, tends to provide better results when evaluating older subjects. These results show that classification on a model trained on G1 subjects will have a good classification for all subjects regardless of their age. In addition, a model trained on older subjects will also produce good results when it comes to classifying older individuals.

We observed that the best model for classification on G1 subjects is trained on G1 subjects. We also know that sleep efficiency decreases with age \cite{moser2009sleep}, affecting model training and classification.

S1 classification improves in models trained with older subjects. Classification performs better when classifying on older subjects in models such as G4-model (Figure \ref{fig: Fig3}). Only G1 subjects decrease classification performance. As before, G4 subjects obtain the best classification of W stage, which improves when we train the model with older subjects. As observed previously in the literature, the number and duration of awakenings increase with age \cite{van2000age}. As we age, an increase in alpha theta power (5-1 Hz) is generated in sleep signals present in the W and S1 stages \cite{moser2009sleep,luca2015age,van2000age}. 

The classification of S2 stage is always better when we use the G1-model, as we can see in Figure \ref{fig: Fig3} and in Table \ref{tab:best_age}. The work of \cite{luca2015age} says the fast spindles present in S2, decrease with age. We suppose that their change could affect the model, being better the model trained with younger subjects, demonstrating the importance of the decrease of spindles in the classification of S2 stage.

On the other hand, SWS has an important decay when we classify over older subjects, which we observed in all models. It can be seen that in younger subjects a better SWS classification was obtained with the G1-model. However, when we want to classify on older subjects, the G4-model obtained better results. This can be explained by the decrease of spectral power densities within the slow waves with increasing age \cite{van2000age,luca2015age}. We therefore assume that, for SWS stage classification, it is better if a model with similar amplitude is trained on the slow-wave stage. 

The G1-model performed best during REM stage classification in all test subjects, except for G4 subjects. For those subjects, the models trained with G2 and G4 performed best. With increasing age, the evening cortisol level rise \cite{van2000age}. This elevation of the cortisol hormone has an inverse relationship with the duration of the REM stage. The longer latency in G1 subjects may improve training in the model, but it may be too different from REM in G4 subjects, with the model trained with the same age group being better.

In the article conducted by Zhou et al. \cite{zhou2020automatic}, they also demonstrated the significance of considering age in sleep stage classification. Their research revealed that incorporating age as a feature led to notable improvements in the overall classification performance when differentiating between various sleep stages. These findings highlight the valuable contribution of age as a relevant factor in enhancing the accuracy and effectiveness of sleep stage classification models.

\subsection{Effect of early-late sleep}

Overall, both models perform better in the early sleep stages. However, the model trained with early sleep performed better. Still, in the classification of late sleep, the model trained with late sleep performed better (Figure \ref{fig: Fig3}). We speculate that this is due to the way in which features are expressed in each sleep half, which means that a specific model trained to learn those features will improve classification \cite{gais2000early}.

In the W stage, we obtained high variability in classifying early sleep compared to classifying late sleep. In this class, the model trained with late sleep performed better than the model trained with early sleep, as we can see in Figure \ref{fig: Fig3}.

In the classification of stage S1, we observe that the model trained with early sleep performed better than the model trained with late sleep. It was shown in \cite{gais2000early} that subjects spend more time in S1 stage in early sleep, although the difference is not significant, we believe this change may improve the model in early sleep.

For stage S2, we observe an improvement when classifying over the late sleep stages, with less variability for each subject. For classifying over early sleep stages, the model trained on early sleep performed better. As with S1 stage, Gains et al. found out that the time in the S2 stage over late sleep is slightly longer than early sleep. This may be the cause for the improvement on late sleep model classification performance  \cite{gais2000early}.

In the SWS stage, we observed differences when classifying over early sleep. This behaviour could be explained by the dominance of SWS present in early sleep \cite{rasch2013sleep,gais2000early}. On the other hand, the REM stage dominates the second half of sleep \cite{rasch2013sleep,gais2000early}. In our experiments, the late sleep model obtained better classification results than the early sleep model.

\subsection{Sleep scoring for sleep disorders}

We observed differences in the classification of sleep stages depending on the model used. An expert might want to analyze more specific disorders, such as REM sleep behaviour \cite{plazzi2008nocturnal}, where a correct classification of the REM stage becomes more important, even at the expense of a decrease in classification performance for other stages. 

For instance, narcoleptic patients present a different sleep pattern, with longer NREM/REM cycles, longer intervals between REM episodes, and an attenuated progression of REM sleep, with good distinction between the REM stage from the others also being important \cite{plazzi2008nocturnal}. A model with higher accuracy in the classification of REM sleep might be of relevance in such a case.

Therefore, the choice of which model is "best", or which data to use for training, should be guided by the study of interest. Important decisions also have to be made depending on the age of the subject and whether we use early or late sleep. 

The code developed during the preparation of this manuscript was uploaded to a git repository on \url{github.com/eugeniaMoris/2022_sleep_scoring_analisis}

\subsection{Limitations}

A limitation of our study was the lack of subjects between 35 and 50 years old in our sample when studying the effects of age, also prescribed by the data-set used. Nevertheless, we use four age groups that covered the most marked differences in sleep quality for healthy adults \cite{luca2015age}. Larger datasets could be incorporated in the future to improve the age distribution.

Another limitation was the significant imbalance present in the data. The data had significantly more samples of W class, while the number of epochs belonging to S1 is comparatively small. This may cause the poor performance in classifying that stage \cite{supratak2017deepsleepnet}. We used RUS to overcome the W stage imbalance. In the S1 class, we do not have a way to expand the number of samples without generating artificial ones. The study can be considered in the future.

Finally, there are two criteria for labelling sleep epochs, the R\&K criteria (used in this work) and the American Academy of Sleep Medicine (AASM) \cite{berry2012aasm} criteria. The data used in this work is pre-labelled following the R\&K criteria. Then, there is no evidence that our model can correctly classify data according to the AASM criterion. In future studies, seeking a method that can generalise between the two criteria might be a way to overcome this issue.


\section{Conclusion}

In this work, the classification of sleep stages was performed with wavelets, as feature extractor; and RF, as a classifier. Model selection, subject age and early-late sleep studies were performed. With the aim of helping in the choice of the best model according to the data to be classified and the problem to be treated. 

We conclude that the use of different types of channels improves our classification by improving the difference between the most similar classes (W, S1, REM). Focusing on these similar classes also improves the models, by using a CM and using statistical variables as features. The former shows that these classes introduce the largest uncertainty and miss-classifications during automatic classification.

We also notice the improvement when we train our model with younger subjects, even for the classification of older subjects. Still, depending on whether we want to focus on better classification for a specific sleep stage, such as the S1 stage, a model trained with older subjects will perform better.

Finally, for early and late sleep classification, a specific model for each half will improve the model performance.

\section*{Acknowledgement(s)}
This work was supported by PICT PICT 2020-00045 granted by Agencia I+D+i (Ministerio de Ciencia, Tecnología e Innovación de la Nación, Argentina) and by PIP 2021-2023 11220200102472CO from CONICET (Argentina).

\section*{Declarations}

\textbf{Funding:}
The research leading to these results received funding from Agencia I+D+i (Ministerio de Ciencia, Tecnología e Innovación de la Nación, Argentina) under Grant Agreement PICT 2020-0045 and by PIP 2021-2023 11220200102472CO from CONICET (Argentina).
The authors declare they have no financial interests.

\textbf{Disclosure statement:}
The authors report there are no competing interests to declare.

\textbf{Ethics approval:}
The data that support the findings of this study are openly available in PhysioNet at \url{https://doi.org/10.13026/C2X676}

\bibliographystyle{spmpsci}      

\bibliography{refs}   

\end{document}